\documentclass[10pt,twocolumn,letterpaper]{article}

\usepackage[pagenumbers]{cvpr} 

\usepackage{graphicx}
\usepackage{amsmath}
\usepackage{amssymb}
\usepackage{booktabs}
\usepackage{algorithm}
\usepackage{algpseudocode}
\usepackage{adjustbox}
\usepackage{pifont}
\usepackage{xcolor}
\usepackage{multirow}

\usepackage[pagebackref,breaklinks,colorlinks]{hyperref}

\usepackage[capitalize]{cleveref}
\crefname{section}{Sec.}{Secs.}
\Crefname{section}{Section}{Sections}
\Crefname{table}{Table}{Tables}
\crefname{table}{Tab.}{Tabs.}

\def\cvprPaperID{10622} 
\def\confName{CVPR}
\def\confYear{2022}

\newcommand{\bbR}{\mathbb{R}}
\newcommand{\vv}{\ding{52}}
\newcommand{\x}{\ding{54}}
\newcommand{\dtw}{\text{DTW}}
\newcommand{\sdtw}{\text{S2DTW}}

\newcommand{\Ac}{\mathcal{A}}

\newcommand{\green}[1]{\textcolor[rgb]{0, 0.5, 0}{#1}}
\newcommand{\blue}[1]{\textcolor[rgb]{0, 0, 1}{#1}}
\newcommand{\red}[1]{\textcolor[rgb]{1, 0, 0}{#1}}

\begin{document}

\title{Video-Text Representation Learning via Differentiable Weak Temporal Alignment}


\newcommand*\samethanks[1][\value{footnote}]{\footnotemark[#1]}
\author{
Dohwan Ko\textsuperscript{\rm 1}\hspace{0.4cm}
Joonmyung Choi\textsuperscript{\rm 1}\hspace{0.4cm}
Juyeon Ko\textsuperscript{\rm 1}\hspace{0.4cm}
Shinyeong Noh\textsuperscript{\rm 1}\vspace{0.0cm} \\
Kyoung-Woon On\textsuperscript{\rm 2}\hspace{0.4cm} 
Eun-Sol Kim\textsuperscript{\rm 3}\hspace{0.4cm}
Hyunwoo J. Kim\textsuperscript{\rm 1}\thanks{is the corresponding author.}\vspace{0.3cm} \\
\textsuperscript{\rm 1}Department of Computer Science and Engineering, Korea University\vspace{0cm} \\
\textsuperscript{\rm 2}Kakao Brain\hspace{0.4cm} \textsuperscript{\rm 3} Department of Computer Science, Hanyang University\vspace{0cm} \\
\tt\small \{ikodoh, pizard, juyon98, dneirfi, hyunwoojkim\}@korea.ac.kr \vspace{0cm}
\\\tt\small \{kloud.ohn\}@kakaobrain.com\hspace{0.4cm} \tt\small \{eunsolkim\}@hanyang.ac.kr \vspace{0cm}
}
\maketitle

\begin{abstract}
Learning generic joint representations for video and text by a supervised method requires a prohibitively substantial amount of manually  annotated video datasets.
As a practical alternative, a large-scale but uncurated and narrated video dataset, HowTo100M, has recently been introduced. 
But it is still challenging to learn joint embeddings of video and text in a self-supervised manner, due to its ambiguity and non-sequential alignment.
In this paper, we propose a novel multi-modal self-supervised framework \textbf{V}ideo-\textbf{T}ext \textbf{T}emporally \textbf{W}eak Al\textbf{i}gnme\textbf{n}t-based Contra\textbf{s}tive Learning (VT-TWINS) to capture significant information from noisy and weakly correlated data using a variant of Dynamic Time Warping (DTW).
We observe that the standard DTW inherently cannot handle weakly correlated data and only considers the globally optimal alignment path. To address these problems, we develop a differentiable DTW which also reflects local information with weak temporal alignment.
Moreover, our proposed model applies a contrastive learning scheme to learn feature representations on weakly correlated data.
Our extensive experiments demonstrate that VT-TWINS attains significant improvements in multi-modal representation learning and outperforms various challenging downstream tasks.
Code is available at \href{https://github.com/mlvlab/VT-TWINS}{https://github.com/mlvlab/VT-TWINS}.
\end{abstract}

\section{Introduction}
\label{sec:intro}

Learning video-text representations is an important problem in computer vision. 
In recent years, it has recently drawn increasing attention due to a large amount of video data and various applications.
Previous works~\cite{lin2014microsoft,zhou2018towards,wray2019fine} have achieved exciting results by learning mappings between video clips and texts
but they usually require a large amount of manual annotations such as MSR-VTT~\cite{xu2016msr}, DiDeMo~\cite{anne2017localizing}, EPIC-KITCHENS~\cite{damen2018scaling}. 
However, since labeling videos is expensive and time-consuming, it does not scale well for sufficiently large datasets which are essential to learning generic video-text representations that are readily applicable to a wide range of downstream tasks that include text-to-video retrieval or video-text retrieval~\cite{klein2015associating,wang2018learning,wang2016learning,yu2018joint}, text-based action localization~\cite{anne2017localizing,cheron2018flexible}, action segmentation~\cite{lea2016temporal,sigurdsson2017asynchronous} and video question answering~\cite{tapaswi2016movieqa,malinowski2015ask,yu2018joint}. 
Recent studies suggest that multi-modal self-supervised learning with a huge amount of data is a promising alternative to fully supervised methods~\cite{fernando2017self,xu2019self}. To this extent, HowTo100M~\cite{miech2019howto100m} has been introduced, which is composed of 100 million pairs of video clips and captions from 1.22M \textit{narrated instructional} videos.

\begin{figure*}[t!]
    \centering
    \begin{subfigure}[b]{0.49\textwidth}  
        \centering 
        \includegraphics[width=\textwidth]{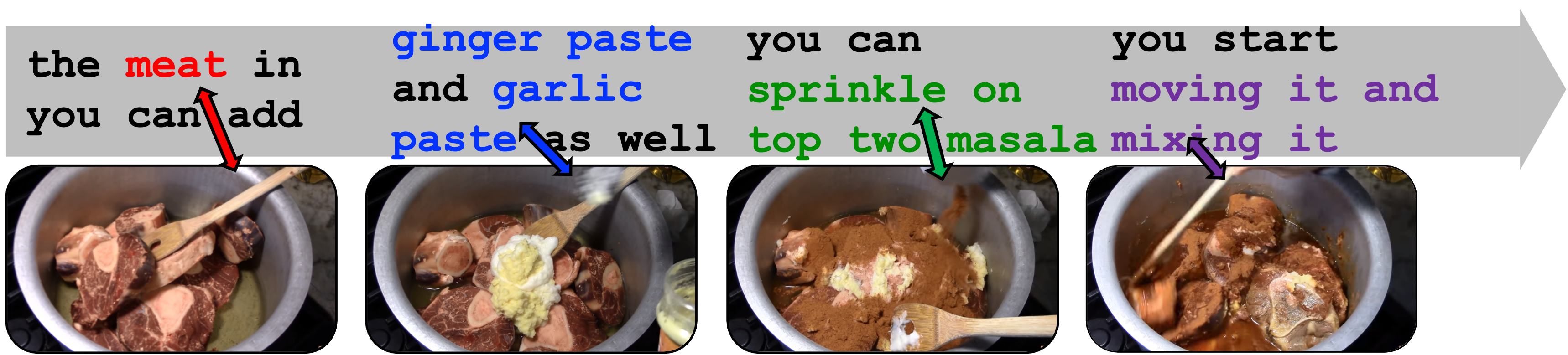}
        \caption{Sequentially aligned pairs}
        \label{fig:sequenital}
    \end{subfigure}
    \hfill
    \begin{subfigure}[b]{0.49\textwidth}
        \centering
        \includegraphics[width=\textwidth]{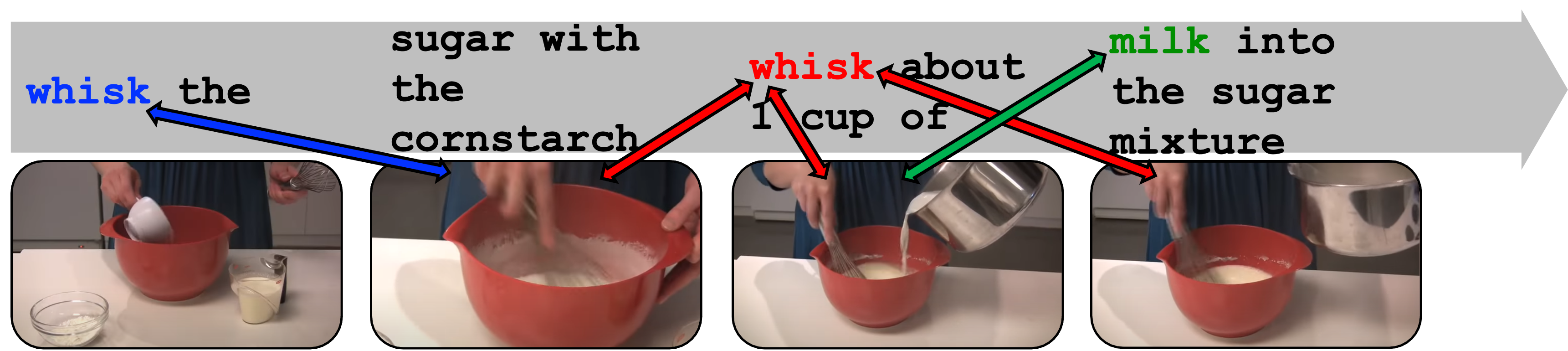}
        \caption{Non-sequentially aligned pairs}
        \label{fig:nonsequenital}
    \end{subfigure}
    \vskip\baselineskip
    \vspace{-0.1cm}
    \begin{subfigure}[b]{0.49\textwidth}   
        \centering 
        \includegraphics[width=\textwidth]{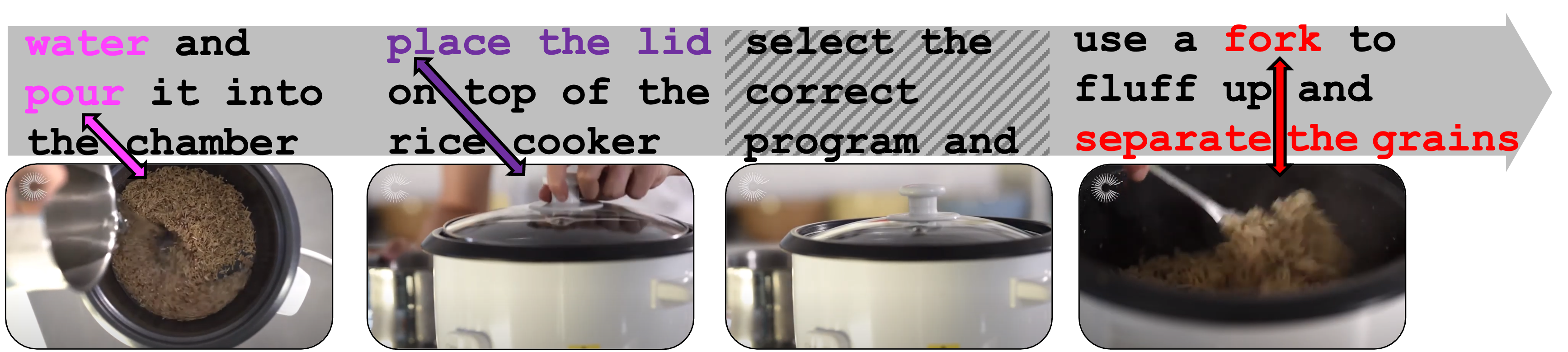}
        \caption{Partially irrelevant pairs}
        \label{fig:partial}
    \end{subfigure}
    \hfill
    \begin{subfigure}[b]{0.49\textwidth}   
        \centering 
        \includegraphics[width=\textwidth]{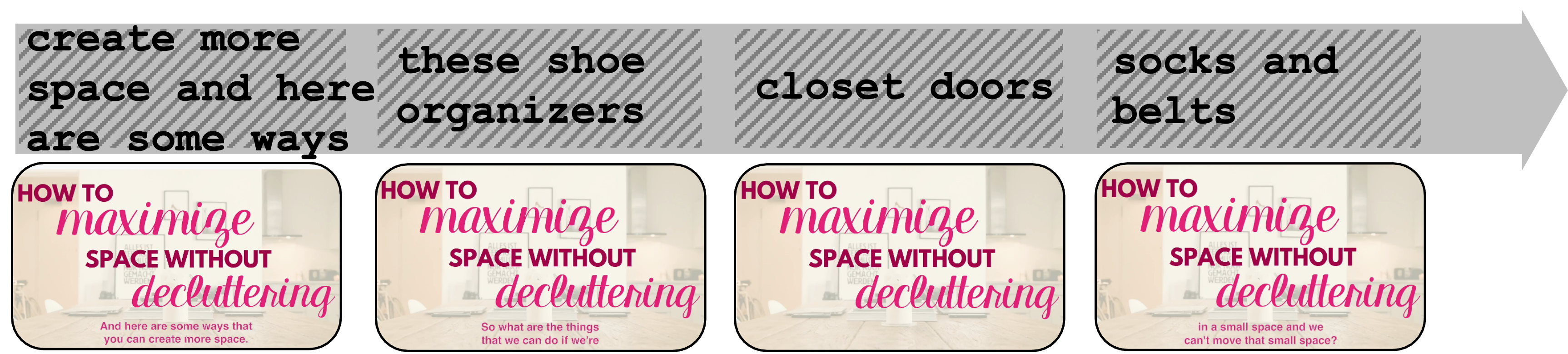}
        \caption{Entirely irrelevant pairs}
        \label{fig:entire}
    \end{subfigure}
    \caption{\textbf{Examples of the HowTo100M.} 
    The HowTo100M dataset contains narrated instructional videos and the clips and captions are weakly aligned.
    Each video is composed of several pairs of clips and captions.
    Shaded captions are irrelevant to any clips in the video.
    }
    \label{fig:howto100m}
\end{figure*}

The HowTo100M is one of the largest video datasets but it comes with several challenges. 
It is uncurated and its video-text pairs are weakly correlated meaning that given a video clip the caption depicting the visual content may appear \textit{before/after} the clip or \textit{not} even exist (Figure~\ref{fig:howto100m}).
To handle the weakly correlated video-text pairs, MIL-NCE~\cite{miech2020end} has proposed a multiple instance learning (MIL)-based contrastive learning adopting Noise Contrastive Learning (NCE) loss \cite{gutmann2010noise}.
MIL-NCE treats the multiple captions which are temporally close to one clip as positive samples allowing one-to-many correspondence.
But this strong assumption often leads to suboptimal representation learning.

In this paper, to address the problem, we develop a new weak temporal alignment algorithm building upon Dynamic Time Warping (DTW)~\cite{sakoe1978dynamic}.
In contrast to the standard DTW which is limited to sequential alignment, our proposed alignment algorithm allows flexibility by skipping irrelevant pairs and starting/ending at arbitrary time points. 
Also, it takes into account a globally optimal path as well as locally optimal paths by introducing local neighborhood smoothing.
More importantly, our alignment algorithm is differentiable so we incorporate it into representation learning as a distance measure.
We then propose a novel multi-modal self-supervised learning framework to learn a joint video and text embedding model named as \textbf{V}ideo-\textbf{T}ext \textbf{T}emporally \textbf{W}eak Al\textbf{i}gnme\textbf{n}t-based Contra\textbf{s}tive Learning (VT-TWINS) that automatically handles the correspondence between noisy and weakly correlated captions and clips.

Our extensive experiments on five benchmark datasets demonstrate that our learned video and text representations generalize well on various downstream tasks including action recognition, text-to-video retrieval, and action step localization.
Moreover, ablation studies and qualitative analysis show that our framework effectively aligns the noisy and weakly correlated multi-modal time-series data.

Our \textbf{contributions} are threefold:
\begin{itemize}
    \item We propose a novel self-supervised learning framework with differentiable weak temporal alignment that automatically handles the noisy and weakly correlated multi-modal time-series data. 
    \item We analyze the local neighborhood smoothing in our alignment algorithm showing that unlike DTW the alignment takes into account local optimal paths as well as global optimal path.
    \item Our experiments show that the proposed method considerably improves joint representations of video and text an is adapted well on various downstream tasks.
\end{itemize}
\section{Related Work}
\label{sec:related}

\noindent \textbf{Self-Supervised Learning for Videos.}
The self-supervised learning approaches have received considerable attention because they do not require additional annotations during learning representation. 
Recently, several works are proposed to learn video representations in a self-supervised manner.
One research direction is to design video-specific pretext tasks, such as verifying temporal orders \cite{lee2017unsupervised, fernando2017self, misra2016shuffle, xu2019self}, predicting video rotation \cite{jing2018self}, solving jigsaw puzzles in a video \cite{kim2019self}, and dense predictive coding \cite{han2019video}. 
Another line of research is to use a contrastive learning which leads clips from the same video to be pulled together while clips from different videos to be pushed away \cite{sun2019learning,wang2020self,qian2021spatiotemporal, chen2020simple, chen2021exploring,grill2020bootstrap,he2020momentum}.
In view of the multi-modality of videos, many works explore mutual supervision across modalities to learn representations of each modality.
For example, they regard temporal or semantic consistency between videos and audios \cite{korbar2018cooperative,chen2021multimodal} or narrations \cite{miech2020end,alayrac2016unsupervised,miech2019howto100m,bain2021frozen} as a natural source of supervision.
MIL-NCE \cite{miech2020end} introduced contrastive learning to learn joint embeddings between clips and captions of unlabeled and uncurated narrated videos.
The other line of work adopts an additional crossmodal encoder (\eg, crossmodal transformer) to capture richer interaction between modalities \cite{sun2019videobert,sun2019learning,zhu2020actbert,li2020hero,luo2020univl,ging2020coot}.
In this paper, we focus on extending contrastive learning to temporally align two time-series modalities, \ie, clips and captions from videos without any additional crossmodal encoders.

\noindent \textbf{Sequence Alignment.}
Sequence alignment is crucial in fields related to the time-series data due to the temporal information. 
In particular, the lack of manually annotated video datasets makes it harder to align clips and captions temporally. 
Dynamic Time Warping (DTW)~\cite{sakoe1978dynamic} measures the distance with strong temporal constraints between two sequences. 
\cite{chang2021learning} uses global sequence alignment as a proxy task by relying on the DTW.
\cite{cuturi2017soft, hadji2021representation} extended the DTW for end-to-end learning with differentiable approximations of the discrete operations (\eg, the `min' operator) in the DTW.
Chang \etal.~\cite{chang2019d3tw} proposed the frame-wise alignment loss using the DTW in weakly supervised action alignment in videos.
Drop-DTW~\cite{dvornik2021drop} proposed a variant of the DTW algorithm which automatically drops the outlier elements from the pairwise distance to handle the noisy data.
However, using the DTW alone can cause feature collapsing which leads all the feature embeddings to be concentrated to a single point.
To address this problem, \cite{chang2019d3tw} and \cite{haresh2021learning} use the subsidiary regularization loss term with the DTW. 

\section{Preliminaries}
\label{sec:preliminaries}
We briefly summarize the basic concepts of dynamic time warping and the characteristics of an uncurated narrated video dataset HowTo100M.

\subsection{Dynamic Time Warping (DTW)}
\label{subsec:dtw}

\textbf{DTW}~\cite{berndt1994using} finds an optimal alignment between two time-series data.
Let $X$ and $Y$ denote two time-series data of length $n$ and $m$, \ie, $X = [x_1, x_2, \dots, x_n]$ and $Y = [y_1, y_2, \dots, y_m]$. 
DTW first computes a pairwise distance matrix $\Delta(X, Y):=[\delta(x_i,y_j)]_{ij} \in \bbR^{n\times m}$ with a distance measure $\delta$.
Then, DTW optimizes the following:
\begin{equation}
    \dtw(X, Y) = \min_{A \in \Ac_{n, m}} \langle A, \Delta(X, Y)\rangle,
    \label{eq:DTW}
\end{equation}
where $\Ac_{n, m} \subset \{0, 1\}^{n \times m}$ is a set of (binary) alignment matrices.
An alignment matrix $A$ represents a path that connects from $(1, 1)$ to $(n , m)$-th entries of $\Delta(X,Y)$ by three possible moves $\{\downarrow, \searrow, \rightarrow\}$.

To efficiently find an optimal path, DTW \cite{berndt1994using} uses dynamic programming to recursively solve the following subproblems:
\begin{equation}
    \label{eq:dynamic}
    r_{i, j} = \delta_{i, j} + \min \{r_{i-1, j}, r_{i, j-1}, r_{i-1, j-1}\},
\end{equation}
where $r_{i, j}$ is the $(i, j)$-th element of a cumulative cost matrix $R(X, Y) \in \bbR^{n\times m}$ of $\Delta(X,Y)$.
Therefore, $\dtw(X, Y)$ in~\eqref{eq:DTW} is equal to $r_{n,m}$ which is the accumulated  cost  that evaluates the similarity between two time-series data.

\textbf{Soft-DTW}~\cite{cuturi2017soft} has proposed a \textit{differentiable} variant of the DTW replacing the non-differentiable operator `$\min$' in ~\eqref{eq:dynamic} with the soft-min `${\min}^\gamma$' defined as:
\begin{equation}
\label{eq:softmin}
{\min}^\gamma \{a_1, a_2, \dots, a_m \} = -\gamma \log \sum_{i=1}^{m} e^{-a_i/\gamma},
\end{equation}
where $\gamma \in \mathbb{R}_{+} $ is a smoothing parameter. 
Then, the recurrence relation of Soft-DTW is given as:
\begin{equation}
    \label{eq:soft_dynamic}
    r_{i, j} = \delta_{i, j} + {\min}^{\gamma} \{r_{i-1, j}, r_{i, j-1}, r_{i-1, j-1}\}.
\end{equation}
If $\gamma$ is zero, soft-min $\min^\gamma$ is identical to $\min$ operator.
As $\gamma$ increases, Soft-DTW(X,Y) more takes into account the cost of suboptimal paths.

\subsection{The HowTo100M Dataset}
\label{subsec:dataset}

HowTo100M dataset~\cite{miech2019howto100m} is a large-scale dataset that contains 136M video clips with paired captions from 1.22M narrated instructional videos across 23K different visual tasks.
A video has 110 clip-caption pairs with an average duration of 4 seconds. 
The captions are automatically transcribed narrations via automatic speech recognition (ASR).
Learning joint video text embeddings with HowTo100M has two sources of difficulties: `uncurated narrations' and `weak correlation' between clip-caption pairs.
As discussed in \cite{miech2020end}, the narrations transcribed by ASR are potentially erroneous and the colloquial language is neither complete nor grammatically correct sentences.
In addition, due to the weak correlation between the paired clips and captions, computing the optimal correspondence to learn joint embedding entails addressing the following challenges, which is the main focus of this paper.

\noindent \textbf{Ambiguity.} 
As aforementioned, the average duration of a clip-caption pair is 4 seconds. 
Since short clips are sampled densely in one video, consecutive clips are often semantically similar, \ie, clip-caption alignments inherently have ambiguity.
So it is more beneficial to use algorithms that take into account multiple alignments allowing many-to-many correspondence rather than the algorithms that consider the only one optimal path such as the standard DTW.

\noindent \textbf{Irrelevant pairs.} 
The paired clips and captions may contain irrelevant contents due to several reasons. 
People might \textit{skip} to demonstrate some steps when narrations are clear enough or vice versa. 
In Figure~\ref{fig:partial}, since the narration ``select the correct program '' is clear enough,  no demonstration is given in the corresponding clip.
In addition, some videos have entirely irrelevant clips and captions like Figure~\ref{fig:entire}.
When learning joint video text embeddings, these irrelevant pairs should be properly handled.

\noindent \textbf{Non-sequential alignment.} 
Although videos and texts are overall correlated at the video-level, the paired clips and captions often are not temporally well-aligned. 
For instance, people in a video describe plans \textit{before} demonstrations or explain details \textit{after} actions, \ie, captions may come with temporal shifts. 
To estimate the correspondence between clips and captions, they can be aligned without changing the order of elements in each modality like Figure~\ref{fig:sequenital}, called \textit{sequential} alignment.
In contrast, when the order of elements in a modality is partially reversed or the content of a clip/caption is arbitrarily interspersed in the other modality, \textit{non-sequential} alignments are required to compute the optimal correspondence.
We observe that the non-sequential alignments often occur when videos have long sequences of captions and clips like Figure~\ref{fig:nonsequenital}.
We will address the challenges by a new learning strategy. 
\begin{figure*}[t] 
    \centering
    \includegraphics[width=1\textwidth]{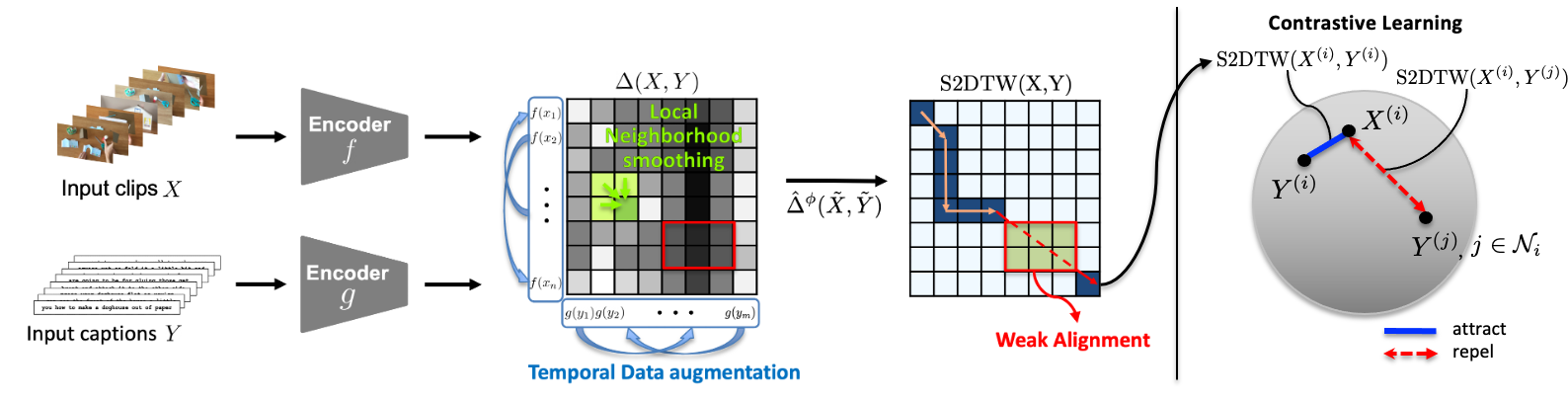}
    \caption{\textbf{Overall Architecture.} 
    We propose a multi-modal self-supervised learning framework, VT-TWINS, to learn joint embeddings of video and text from noisy and weakly correlated data.
    The encoders $f$ and $g$ firstly extract feature embeddings from input clips $X$ and input captions $Y$, respectively. 
    Then, we present a new alignment algorithm based on the DTW, called S2DTW, which can handle weakly correlated data with local neighborhood smoothing (Section~\ref{subsec:smoothing}) and weak alignment (Section~\ref{subsec:skip}). 
    We also apply temporal data augmentation (Section~\ref{subsec:augmentation}) to learn from the non-sequentially aligned data with the S2DTW.
    We finally employ a contrastive learning scheme, which uses the S2DTW as a distance measure between the clip-caption pairs, with negative pairs ($\mathcal{N}_i$) for representation learning while preventing feature collapsing (Section~\ref{subsec:negative}). 
    }
    \label{fig:overall}
\end{figure*}
\section{Method}
\label{sec:method}
In this section, we present a novel multi-modal self-supervised framework, named as \textbf{V}ideo-\textbf{T}ext \textbf{T}emporally \textbf{W}eak Al\textbf{i}gnme\textbf{n}t-based Contra\textbf{s}tive Learning (VT-TWINS), to learn joint embeddings of video and text from uncurated narrated videos.
To address the problems mentioned above and estimate more accurate correspondence, we propose a new differentiable variant of DTW, called Locally \textbf{S}moothed \textbf{S}oft-\textbf{DT}W with \textbf{W}eak Alignment (S2DTW). First, we apply local neighborhood smoothing and weak alignment.
We then adopt temporal data augmentation for non-sequential alignments that the standard DTW cannot inherently handle.
We finally apply a contrastive learning scheme and present VT-TWINS for representation learning without feature collapsing.
Figure~\ref{fig:overall} and Algorithm~\ref{alg:overall} show our overall algorithm VT-TWINS including S2DTW.

\subsection{Local Neighborhood Smoothing}
\label{subsec:smoothing}

To address the ambiguity as mentioned in Section~\ref{subsec:dataset}, we smooth the pairwise distance matrix $\Delta(X, Y)$ as:
\begin{equation}
    \label{eq:smoothing}
    \hat{\delta}_{i, j} = \delta_{i, j} + {\min}^\gamma \{\delta_{i-1, j}, \delta_{i, j-1}, \delta_{i-1, j-1}\},
\end{equation}
where $\delta_{i, j}$ and $\hat{\delta}_{i, j}$ are the $(i, j)$-th elements of $\Delta(X, Y)$ and $\hat{\Delta}(X, Y)$, respectively.
This allows many-to-many correspondence and encourages the alignment algorithm to focus more on a locally optimal clip (or caption), which has relatively smaller distances to others within a small neighborhood.
$\hat{\delta}_{i, j}$ can be viewed as smoothed $\delta_{i, j}$ with its previous elements $\delta_{i-1, j}$, $\delta_{i, j-1}$, and $\delta_{i-1, j-1}$.
Then, similar to \eqref{eq:soft_dynamic} we apply dynamic programming to compute the optimal cost from smoothed distance matrix $\hat{\Delta}(X, Y)$ instead of $\Delta(X, Y)$ and as follows: 
\begin{equation}
    \label{eq:cumulative}
    \hat{r}_{i, j} = \hat{\delta}_{i, j} + {\min}^\gamma \{\hat{r}_{i-1, j}, \hat{r}_{i, j-1}, \hat{r}_{i-1, j-1}\}.
\end{equation}
S2DTW decays the cost of older matches and reflects more recent elements since \eqref{eq:cumulative} accumulates the cost from the top-left element to the bottom-right element, sequentially. 
Roughly speaking, the proposed S2DTW with $\hat{\Delta}(X, Y)$ considers local optimality by \eqref{eq:smoothing} as well as global optimality by \eqref{eq:cumulative} since S2DTW can be rewritten as:
\begin{equation}
\begin{split}
\hat{r}_{i, j} &= \delta_{i, j} + {\min}^\gamma \{\hat{r}_{i-1, j}, \hat{r}_{i, j-1}, \hat{r}_{i-1, j-1}\} \\
&+ {\min}^\gamma \{\delta_{i-1, j}, \delta_{i, j-1}, \delta_{i-1, j-1}\}.
\end{split}
\end{equation}

\begin{figure}[!t] 
    \centering
    \includegraphics[width=0.47\textwidth]{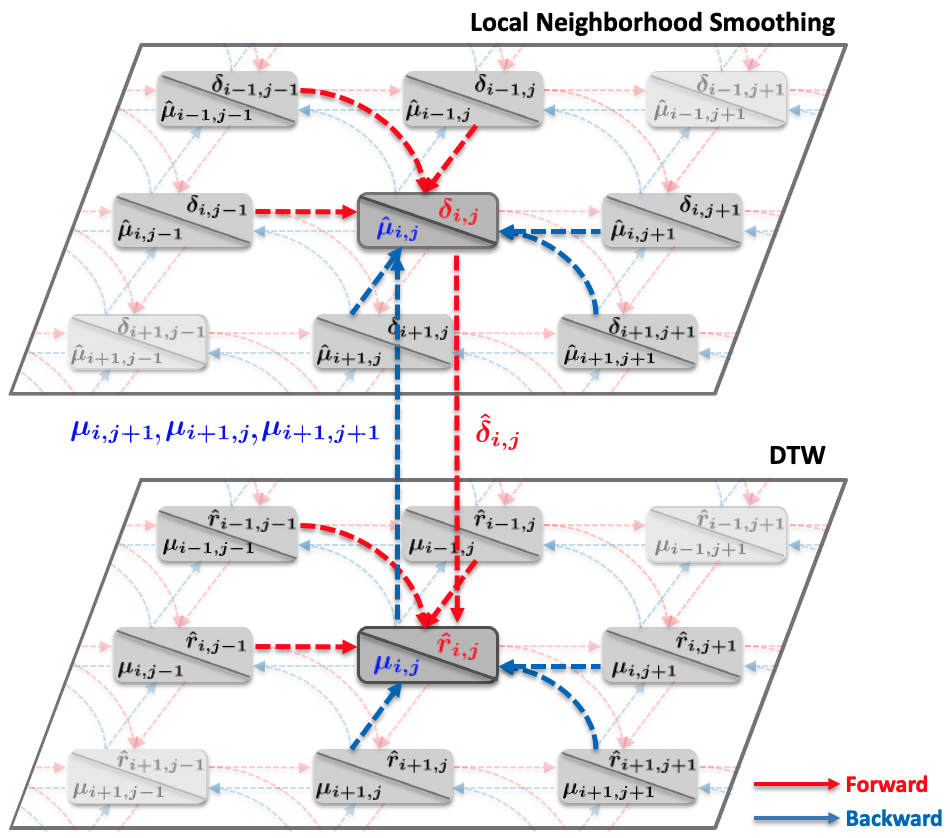}
    \caption{\textbf{Forward and Backward of Local Neighborhood Smoothing in terms of $(i, j)$.}
    At forward propagation, we firstly compute $\hat{\delta}_{i, j}$ by smoothing $\delta_{i, j}$ with $\delta_{i-1, j}$, $\delta_{i, j-1}$, and $\delta_{i-1, j-1}$ in \eqref{eq:smoothing}. 
    Then, $\hat{r}_{i, j}$ is calculated with smoothed $\hat{\delta}_{i, j}$ and $\hat{r}_{i-1, j}$, $\hat{r}_{i, j-1}, \hat{r}_{i-1, j-1}$ by \eqref{eq:cumulative}. 
    At backward propagation, $\mu_{i, j}$ is calculated by \eqref{eq:ours_chain}. 
    It gains the gradient from three directions proportional to how optimal the cumulative cost $\hat{r}$ of each direction is.
    Then, $\hat{\mu}_{i, j}$ is calculated from $\mu_{i+1, j}$, $\mu_{i, j+1}$, and $\mu_{i+1, j+1}$ proportional to how optimal each pairwise cost $\delta$ is in \eqref{eq:smoothing_chain}.
    }
    \label{fig:smoothing}
\end{figure}

\noindent\textbf{Differentiation.} 
We compare Soft-DTW~\cite{cuturi2017soft} and S2DTW via their derivatives.
At the Soft-DTW, they denote a gradient matrix $M= [\mu_{i, j}]$ where $\mu_{i, j} := \frac{\partial r_{n,m}}{\partial \delta_{i, j}} = \frac{\partial r_{n,m}}{\partial r_{i, j}} \cdot \frac{\partial r_{i, j}}{\partial \delta_{i, j}} = \frac{\partial r_{n,m}}{\partial r_{i, j}} \cdot 1 = \frac{\partial r_{n,m}}{\partial r_{i, j}}$ by differentiating \eqref{eq:soft_dynamic} w.r.t $\delta_{i, j}$.
In S2DTW case, however, $\frac{\partial \hat{r}_{n, m}}{\partial \delta_{i, j}} \neq \frac{\partial \hat{r}_{n, m}}{\partial \hat{r}_{i, j}}$ due to the local neighborhood smoothing layer, \ie,  $\frac{\partial \hat{r}_{i, j}}{\partial \delta_{i, j}} \neq 1$.
We therefore redefine $\mu_{i, j} := \frac{\partial \hat{r}_{n, m}}{\partial \hat{r}_{i, j}}$ and denote additional $\hat{\mu}_{i, j} := \frac{\partial \hat{r}_{n, m}}{\partial \delta_{i, j}}$ for the gradient matrix for local neighborhood smoothing layer.
$\mu_{i,j}$ of S2DTW is calculated as follows:
\begin{equation}
    \label{eq:ours_chain}
    \resizebox{1.0\hsize}{!}{$\underbrace{\frac{\partial \hat{r}_{n,m}}{\partial \hat{r}_{i, j}}}_{\mu_{i, j}} = \underbrace{\frac{\partial \hat{r}_{n,m}}{\partial \hat{r}_{i+1, j}}}_{\mu_{i+1, j}} \cdot \green{\dfrac{\partial \hat{r}_{i+1, j}}{\partial \hat{r}_{i, j}}} + \underbrace{\frac{\partial \hat{r}_{n,m}}{\partial \hat{r}_{i, j+1}}}_{\mu_{i, j+1}} \cdot \blue{\dfrac{\partial \hat{r}_{i, j+1}}{\partial \hat{r}_{i, j}}} \\
    + \underbrace{\frac{\partial \hat{r}_{n,m}}{\partial \hat{r}_{i+1, j+1}}}_{\mu_{i+1, j+1}} \cdot \red{\dfrac{\partial \hat{r}_{i+1, j+1}}{\partial \hat{r}_{i, j}}}.$}
\end{equation}
By differentiating \eqref{eq:cumulative} with $i+1$ instead of $i$, the green term of \eqref{eq:ours_chain} is calculated as:
\begin{equation}
    \label{eq:ours_green}
    \gamma \log \green{\frac{\partial \hat{r}_{i+1, j}}{\partial \hat{r}_{i, j}}} = {\min}^\gamma \{\hat{r}_{i, j-1}, \hat{r}_{i, j}, \hat{r}_{i+1, j-1}\} - \hat{r}_{i, j}.
\end{equation}
After calculating $\mu_{i, j}$ in \eqref{eq:ours_chain}, $\hat{\mu}_{i, j}$ is calculated as:
\begin{equation}
    \label{eq:smoothing_chain}
    \resizebox{1.0\hsize}{!}{$\underbrace{\frac{\partial \hat{r}_{n,m}}{\partial \delta_{i, j}}}_{\hat{\mu}_{i, j}} = \underbrace{\frac{\partial \hat{r}_{n,m}}{\partial \hat{r}_{i+1, j}}}_{\mu_{i+1, j}} \cdot \green{\dfrac{\partial \hat{r}_{i+1, j}}{\partial \delta_{i, j}}} + \underbrace{\frac{\partial \hat{r}_{n,m}}{\partial \hat{r}_{i, j+1}}}_{\mu_{i, j+1}} \cdot \blue{\dfrac{\partial \hat{r}_{i, j+1}}{\partial \delta_{i, j}}} \\
    + \underbrace{\frac{\partial \hat{r}_{n,m}}{\partial \hat{r}_{i+1, j+1}}}_{\mu_{i+1, j+1}} \cdot \red{\dfrac{\partial \hat{r}_{i+1, j+1}}{\partial \delta_{i, j}}}.$}
\end{equation}
In \eqref{eq:smoothing_chain}, $\green{\frac{\partial \hat{r}_{i+1, j}}{\partial \delta_{i, j}}} = \frac{\partial \hat{r}_{i+1, j}}{\partial \hat{\delta}_{i+1, j}} \cdot \frac{\partial \hat{\delta}_{i+1, j}}{\partial \delta_{i, j}} = \frac{\partial \hat{\delta}_{i+1, j}}{\partial \delta_{i, j}}$ since $\frac{\partial \hat{r}_{i+1, j}}{\partial \hat{\delta}_{i+1, j}} = 1$.
Similar to \eqref{eq:ours_green}, it is written as:
\begin{equation}
    \label{eq:smoothing_green}
    \gamma \log \green{\frac{\partial \hat{r}_{i+1, j}}{\partial \delta_{i, j}}} = {\min}^\gamma \{\delta_{i, j-1}, \delta_{i, j}, \delta_{i+1, j-1}\} - \delta_{i, j},
\end{equation}
and the other blue and red terms are calculated in the same way.
Like \eqref{eq:ours_green} which measures how minimal the $\hat{r}_{i, j}$ is among three directions, \eqref{eq:smoothing_green} measures how minimal the $\delta_{i, j}$ is among three directions.
Hence, \eqref{eq:ours_chain} aggregates global optimal path information and \eqref{eq:smoothing_chain} aggregates local optimal path information due to the $\hat{r}_{i, j}$ at the former one and the $\delta_{i, j}$ at the latter one.
Unlike S2DTW, the Soft-DTW only requires to calculate $M$ matrix with $r$ instead of $\hat{r}$ by \eqref{eq:ours_chain} and then does not consider the local optimality.
Figure~\ref{fig:smoothing} depicts the forward and backward propagation of S2DTW.

\begin{algorithm}[t!]
\small{
  \caption{VT-TWINS Algorithm with S2DTW}
  \label{alg:overall}
  \textbf{Inputs:} \text{clips} $X$, \text{captions} $Y$ \\
\textbf{Parameters:} smoothing parameter $\gamma$, dummy elements $\phi$ \\
\begin{algorithmic}[1]
    \item[] \textit{\green{\# Temporal Data Augmentation}}
    \State $\forall i,\forall j \in \text{mini-batch} \newline
    \tilde{X}^{(i)},\tilde{Y}^{(j)} \leftarrow Aug(X^{(i)}), Aug(Y^{(j)})$

    \item[] \textit{\green{\# Apply Contrastive Learning Scheme}}
    \State {$\mathcal{L} \leftarrow -\log\sum\limits_{i}\left(\frac{e^{-\sdtw\left(\tilde{X}^{(i)}, \tilde{Y}^{(i)}\right)}}{e^{-\sdtw\left(\tilde{X}^{(i)}, \tilde{Y}^{(i)}\right)} + \sum\limits_{j \in \mathcal{N}_i}e^{-\sdtw\left(\tilde{X}^{(i)}, \tilde{Y}^{(j)}\right)}}\right)$}
    
    \item[] \textit{\green{\# S2DTW}}
    \Function{S2DTW}{${X},{Y}$} 
        \State $\delta_{i, j} \leftarrow \Delta({X},{Y})[i, j], \:\: \forall i \in [1, n], \forall j \in [1, m]$ 
        \item[] \quad \:\:\textit{\green{\# Local Neighborhood Smoothing}}
        \For{$(i, j) = (1, 1)$ {\bfseries to} $(n, m)$}
            \State $\hat{\delta}_{i, j} \leftarrow \delta_{i, j} + {\min}^{\gamma}\{\delta_{i-1, j}, \delta_{i, j-1}, \delta_{i-1, j-1}\}$
        \EndFor 
        \item[] \quad\:\:\textit{\green{\# Weak Alignment}}
        \State $\hat{\Delta}^\phi \leftarrow \text {merge ($\hat\Delta$, $\phi$ )}$
        \item[] \quad\:\:\textit{\green{\# Calculate DTW}}
        \For{$(i, j) = (1, 1)$ {\bfseries to} $(n, m)$}
            \State $\hat{r}_{i, j} \leftarrow \hat{\delta}_{i, j} + {\min}^{\gamma}\{\hat{r}_{i-1, j}, \hat{r}_{i, j-1}, \hat{r}_{i-1, j-1}\}$
        \EndFor  \\
        \quad\: \Return $\hat{r}_{n, m}$
    \EndFunction
    
\end{algorithmic}
\textbf{Output:} $\mathcal{L}$
}
\end{algorithm}
\subsection{Weak Alignment}
\label{subsec:skip}

We further modify the Soft-DTW by allowing its path not to forcibly align irrelevant pairs as (Figure~\ref{fig:partial} and \ref{fig:entire}). 
Besides, our S2DTW can start from (or end at) an arbitrary point.
Adopting the trick in DWSA~\cite{shen2021learning} for one-to-one matching with skipping, we achieve weak alignment by inserting dummy elements $\phi$ in the intervals (and both ends) of clip and caption sequences, (\eg, $X = [x_1, x_2, \dots , x_n]$ becomes $X^\phi = [\phi, x_1, \phi, x_2, \phi, \dots, \phi, x_n, \phi]$).

\begin{figure}
    \begin{subfigure}[b]{0.23\textwidth}   
        \centering 
        \includegraphics[width=\textwidth]{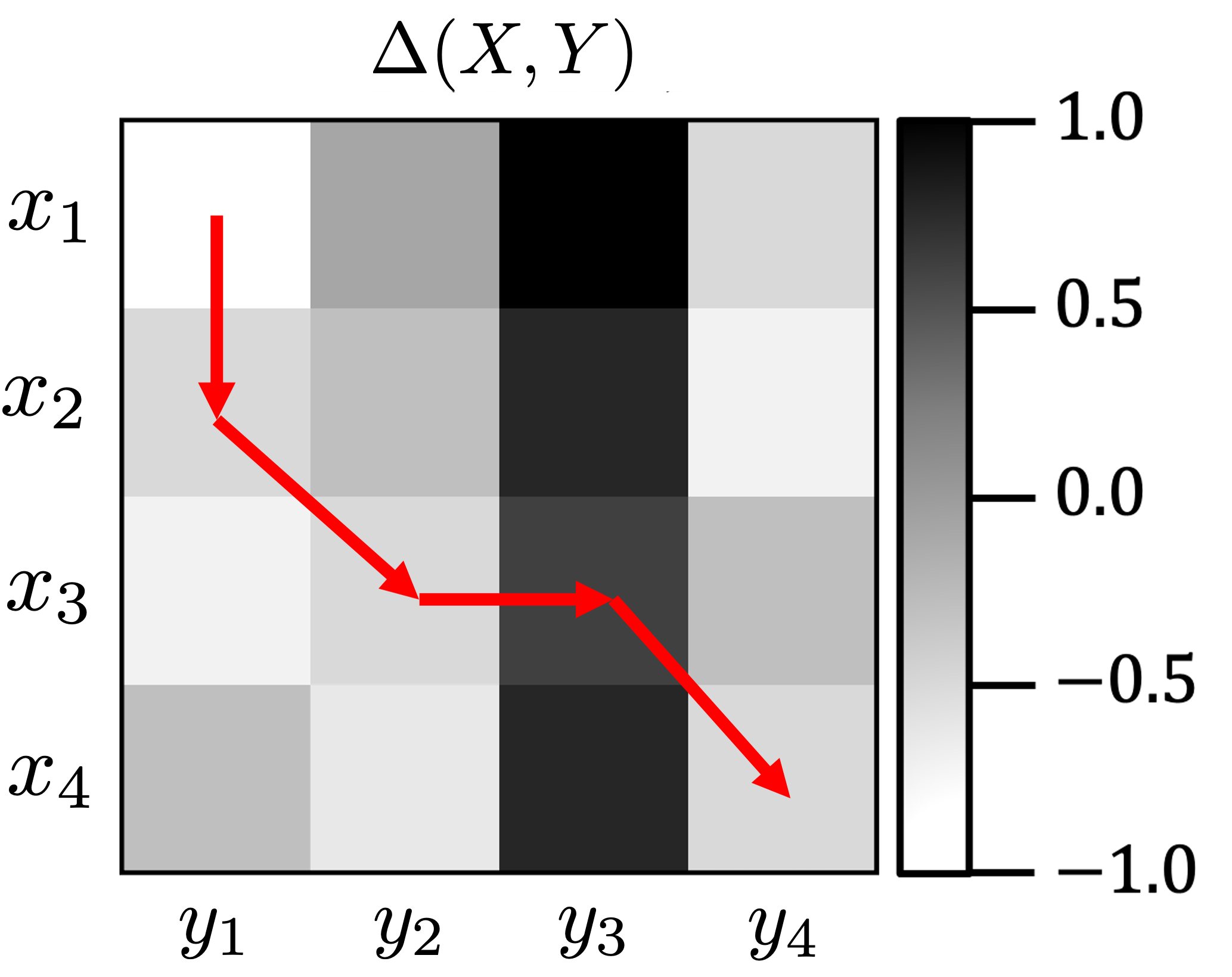}
        \caption{Original Pair-wise Distance}
        \label{fig:original_distance}
    \end{subfigure}
    \hfill
    \begin{subfigure}[b]{0.23\textwidth}   
        \centering 
        \includegraphics[width=\textwidth]{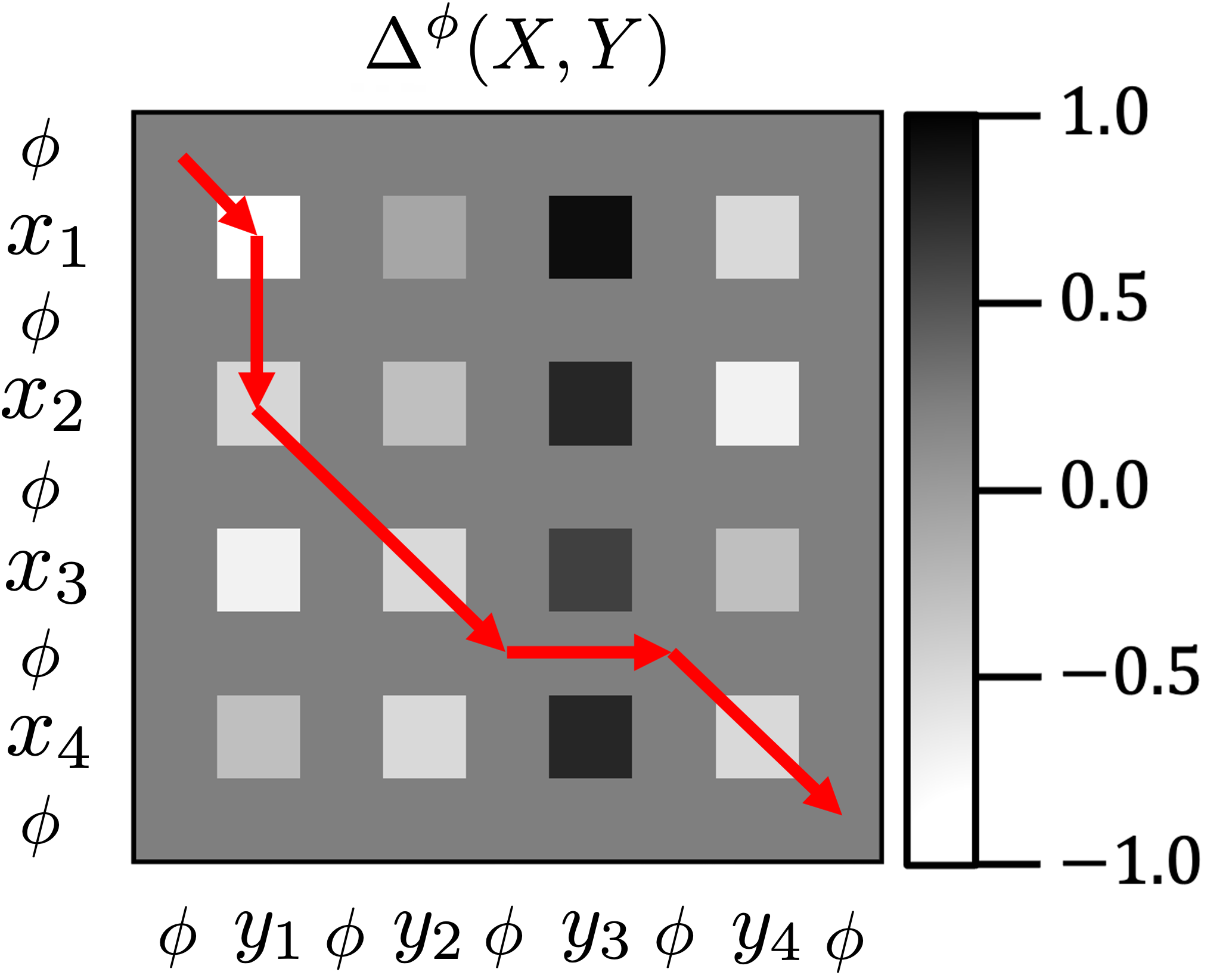}
        \caption{Pair-wise Distance with Dummy}
         \label{fig:dummy_distance}
    \end{subfigure}
    \caption{\textbf{Illustration of Weak Alignment.} 
    (a): The original pairwise distance without dummy elements has to pass the pair whose clip and caption are irrelevant each other, \eg, the caption $y_3$ is irrelevant with any other clips. 
    (b): On the other hand, the pair whose distance is bigger than dummy distance can be skipped.
    }
\end{figure}

In S2DTW, the pairwise distance matrix with dummy elements is $\Delta^{\phi}(X, Y) \in \bbR^{(2n+1) \times (2m+1)}$ and has dummy distance $\delta^\phi$ at the pair which includes $\phi$.
$\delta^{\phi}$ is a hyperparameter that can be interpreted as a threshold.
By calculating the DTW with dummy elements, it leads the DTW path to pass only the pair whose distance is smaller than $\delta^{\phi}$.
Unlike the standard DTW or Soft-DTW which forcibly align at least one pair per one timestamp, our proposed S2DTW weakly aligns the irrelevant clip-caption pairs and even enable many-to-many matchings which cannot be handled by DWSA.
Figure~\ref{fig:original_distance} and \ref{fig:dummy_distance} show the pairwise distance before/after adding dummy elements.
This weak alignment framework is followed by the local neighborhood smoothing. As a result, the final pairwise distance is $\hat{\Delta}^{\phi}(X, Y)$ which is used to calculate the DTW.

\subsection{Temporal Data Augmentation}
\label{subsec:augmentation}
As discussed in Section~\ref{subsec:dataset}, videos often have non-sequential alignments, but the standard DTW cannot resolve them since it allows only three moves $\{\downarrow, \searrow, \rightarrow\}$.
To address this problem, we propose a simple data augmentation that temporally shuffles clips and captions.
Let $\pi$ denote a permutation and then a clip permuted by $\pi$ is $X_{\pi} = [x_{\pi(1)}, x_{\pi(2)}, \dots, x_{\pi(n)}]$.
To avoid temporally or semantically too extreme augmentations, we consider a subset of possible permutations.
We first leave out the cases when a clip is temporally shifted beyond a time window. 
For example of $\forall j \in [1, n]$, the $j$-th clip cannot be out of the window of size $w$, \ie, the range of possible indices after a permutation of $j$-th clip is $[\max(1, j - w), \min(n, j + w)]$. 
The set of permutations that satisfies this temporal constraint is denoted as $\mathcal{T}(n,w)$.
Given the temporal constraint, we propose the target distribution as follows:
\begin{equation}
    \label{eq:probability}
    p^{clip}_{\pi} = 
    \begin{cases} 
        \sigma \left(-\frac{\|\Delta(X, X) - \Delta(X_{\pi}, X_{\pi})\|_2^2}{\tau} \right) & \text{if $\pi \in \mathcal{T}(w,n)$} \\ 
        0 & \text{otherwise}
    \end{cases},
\end{equation}
where $\sigma$ is softmax function computed over all permutations in $\mathcal{T}(w,n)$ and $\tau$ is a temperature parameter. 
$\Delta(X,X)$ and $\Delta(X_{\pi},X_{\pi})$ are self-similarity matrices before/after permutation.
The proposed target distribution more likely generates a permutation that less changes the self-similarity structure. In other words, the proposed augmentation less likely generates semantically too strong augmentations that hinder representation learning.
Then, the temporally augmented $\tilde{X} \sim P^{clip}(X; \Pi)$ which is a shuffled sequence of clips is sampled from the distribution $P^{clip}$ defined in \eqref{eq:probability}.
The captions $\tilde{Y}$ is augmented in the same way and finally we calculate the pairwise distance matrix $\Delta(\tilde{X},\tilde{Y})$ as the input for alignment (\eg, DTW). For simplicity of implementation, each modality is shuffled independently. 

Our temporal augmentation encourages learning invariant features under permutation and allow minimizing the distance between clips and captions that cannot be aligned by sequential alignment algorithms such as the standard DTW.
This is helpful to learn representation when the clips and captions are non-sequentially aligned as in Figure~\ref{fig:nonsequenital}.

\subsection{Contrastive Learning with S2DTW}
\label{subsec:negative}
With S2DTW, we perform representation learning in a self-supervised manner. 
S2DTW can be used for a distance measure between clips and captions. 
Minimizing the distance between two samples without negative pairs causes feature collapsing. 
Hence, to address this problem, we adopt a well-known contrastive loss, InfoNCE loss~\cite{oord2018representation}. 
Our final loss is defined as:
\begin{equation}
    \label{eq:final}
    \resizebox{1.0\hsize}{!}{$\mathcal{L} = -\log\sum\limits_{i}\left(\frac{e^{-\sdtw\left(X^{(i)}, Y^{(i)}\right)}}{e^{-\sdtw\left(X^{(i)}, Y^{(i)}\right)} + \sum\limits_{j \in \mathcal{N}_i}e^{-\sdtw\left(X^{(i)}, Y^{(j)}\right)}}\right)$}
\end{equation}
, where $X^{(i)}$ and $Y^{(i)}$ are clips and captions of the $i$-th video and $\mathcal{N}_i$ is a set of negative samples of the $i$-th video in mini-batch.
This formulation also implicitly mines the hard negatives.
In a clip-caption level, due to the nature of the DTW, a clip-caption pair which has closer distance in negative samples will get stronger negative signal to push away than others in negative samples. 
Therefore, unlike in baseline~\cite{kalantidis2020hard}, no additional hard negative mining strategy (\eg,~\cite{he2020momentum}) was taken for proposed method.
Further discussions with qualitative results are in the appendix.
\section{Experiments}
\label{sec:experiments}
In this section, we evaluate the performance on various downstream tasks by applying our pretrained feature embeddings (Section~\ref{subsec:downstream}).
We also describe ablation studies about the effect of each algorithm which is addressed in Section~\ref{sec:method} and finally analyze qualitative results of each algorithm in terms of the DTW path (Section~\ref{subsec:ablation_qualitative}).
All downstream tasks and ablation studies except for the action recognition task are conducted in the zero-shot learning setting to evaluate only the quality of learned representations.
For the action recognition task, we adopt widely used linear evaluation protocol, which trains a linear classifier on top of the frozen representation.
The experimental setup and further ablation studies are in the appendix.

\subsection{Downstream Tasks}
\label{subsec:downstream}

\subsubsection{Action Recognition}
We firstly evaluate learned video representation without using text representation on the action recognition task whose goal is to distinguish video-level actions.
In Table~\ref{tab:action}, we compare the proposed method with other self-supervised methods. 
According to the linear evaluation protocol, our VT-TWINS outperforms all self-supervised learning methods including the baselines that performed fine-tuning denoted by (Frozen x) such as CBT~\cite{sun2019learning} and 3DRotNet~\cite{jing2018self}.
This result shows that our method improves the generality of video representations. 
Especially for HMDB, VT-TWINS obtains about 4\% improvement over the MIL-NCE with the same backbone model (S3D).

\subsubsection{Video and Text Retrieval}
We evaluate the effectiveness of the joint representation of video and text by applying text-to-video and video-to-text retrieval tasks, which aim to find a corresponding clip (caption) given a query caption (clip).

\begin{table}[t!]
    \centering
    \setlength{\tabcolsep}{3.5pt}
    \begin{adjustbox}{width=0.47\textwidth}
    \begin{tabular}{l|l|c|c|c|cc}
        \toprule
        \multicolumn{1}{c|}{\textbf{Method}} & \multicolumn{1}{c|}{\textbf{Dataset}} & \multicolumn{1}{c|}{\textbf{MM}} & \multicolumn{1}{c|}{\textbf{Model}} & \multicolumn{1}{c|}{\textbf{Frozen}} & \multicolumn{1}{c}{\textbf{HMDB}} & \multicolumn{1}{c}{\textbf{UCF}} \\
        \midrule
        \midrule
        OPN~\cite{alayrac2016unsupervised} & UCF & \x & VGG & \x & 23.8 & 59.6 \\
        Shuffle \& Learn~\cite{misra2016shuffle}* & K600 & \x & S3D & \x & 35.8 & 68.7 \\ 
        Wang \etal~\cite{wang2019self} & K400 & Flow & C3D & \x & 33.4 & 61.2 \\
        CMC~\cite{tian2020contrastive} & UCF & Flow & CaffeNet & \x & 26.7 & 59.1 \\
        Geometry~\cite{gan2018geometry} & UCF & Flow & CaffeNet & \x & 26.7 & 59.1 \\
        Fernanado \etal~\cite{fernando2017self} & UCF & \x & AlexNet & \x & 32.5 & 60.3 \\
        ClipOrder~\cite{xu2019self} & UCF & \x & R(2+1)D & \x & 30.9 & 72.4 \\
        3DRotNet~\cite{jing2018self}* & K600 & \x & S3D & \x & 40.0 & 75.3 \\
        DPC~\cite{han2019video} & K400 & \x & 3D-R34 & \x & 35.7 & 75.7 \\
        3D ST-puzzle~\cite{kim2019self} & K400 & \x & 3D-R18 & \x & 33.7 & 65.8 \\
        CBT~\cite{sun2019learning} & K600 & \x & S3D & \vv & 29.5 & 54.0 \\
        CBT~\cite{sun2019learning} & K600 & \x & S3D & \x & 44.6 & 79.5 \\
        AVTS~\cite{korbar2018cooperative} & K600 & Audio & I3D & \x & 53.0 & 83.7 \\
        MIL-NCE~\cite{miech2020end} & HTM & Text & I3D & \vv & 54.8 & 83.4 \\
        MIL-NCE~\cite{miech2020end} & HTM & Text & S3D & \vv & 53.1 & 82.7 \\
        \midrule
        VT-TWINS & HTM & Text & S3D & \vv & \textbf{57.9} & \textbf{85} \\
        \midrule
        \multicolumn{3}{c|}{\textcolor{gray}{S3D (supervised learning)~\cite{xie2018rethinking}}} & \textcolor{gray}{S3D} & \x & \textcolor{gray}{75.9} & \textcolor{gray}{96.8} \\
        \bottomrule
    \end{tabular}
    \end{adjustbox}
    \caption{\textbf{Action Recognition.}
    Shuffle \& Learn* and 3DRotNet* are reimplemented by~\cite{sun2019learning} with S3D.
    }
    \label{tab:action}
\end{table}
\begin{table}[t!]
    \centering
    \setlength{\tabcolsep}{3.5pt}
    \begin{adjustbox}{width=0.47\textwidth}
    \begin{tabular}{l|l|c c c c}
        \toprule
        \multicolumn{1}{c|}{\textbf{Method}} & \multicolumn{1}{c|}{\textbf{Labeled Dataset}} & R@1 & R@5 & R@10 & MedR \\
        \midrule
        \midrule
        Random Init & None & 0.03 & 0.15 & 0.3& 1675 \\
        HGLMM FC CCA~\cite{klein2015associating} & IM, K400, YC2 & 4.6 & 14.3 & 21.6 & 75 \\
        Miech \etal~\cite{miech2019howto100m} & IM, K400 & 6.1 & 17.3 & 24.8 & 46 \\
        Miech \etal~\cite{miech2019howto100m} & IM, K400, YC2 & 8.2 & 24.5 & 35.3 & 24 \\
        COOT~\cite{ging2020coot} & YC2 & 5.9 & 16.7 & 24.8 & 49.7 \\
        ActBERT~\cite{zhu2020actbert} & YC2 & 9.6 & 26.7 & 38.0 & 19 \\ 
        MIL-NCE~\cite{miech2020end} & None & 8.8 & 24.3	& 34.6 & 23 \\ 
        \midrule
        VT-TWINS & None & \textbf{9.7} & \textbf{27} & \textbf{38.8} & \textbf{19} \\
        \bottomrule
    \end{tabular}
    \end{adjustbox}
    \caption{\textbf{Text-to-Video Retrieval on YouCook2.}}
    \label{tab:text2video_youcook}
\end{table}

\noindent\textbf{Text-to-video retrieval.} Table~\ref{tab:text2video_youcook} and~\ref{tab:text2video_msrvtt} show the performance of text-to-video retrieval on YouCook2 and MSR-VTT dataset.
For fair comparison with MIL-NCE, we trained our model on HowTo100M dataset and evaluate on the test set \textit{without} any additional supervision.
Table~\ref{tab:text2video_youcook} shows that our VT-TWINS outperforms MIL-NCE and even other methods (\eg, COOT and ActBERT) that are fine-tuned on YouCook2 (denoted as YC2).
Similarly, on MSR-VTT dataset Table ~\ref{tab:text2video_msrvtt} shows that the proposed method outperforms several multi-modal self-supervised methods trained on the HowTo100M (MIL-NCE, Amrani \etal., SSB).
In addition, our method is better or on par with ActBert that is fine-tuned on the target dataset MSR-VTT.

\begin{table}[t!]
    \centering
    \setlength{\tabcolsep}{3.5pt}
    \begin{adjustbox}{width=0.47\textwidth}
    \begin{tabular}{l|l|c c c c}
        \toprule
        \multicolumn{1}{c|}{\textbf{Method}} & \multicolumn{1}{c|}{\textbf{Labeled Dataset}} & R@1 & R@5 & R@10 & MedR \\
        \midrule
        \midrule
        Random Init & None & 0.01 & 0.05 & 0.1 & 500 \\
        Miech \etal~\cite{miech2019howto100m} & IM, K400 & 7.5 & 21.2 & 29.6 & 38 \\
        Amrani \etal~\cite{amrani2020noise} & None & 8.0 & 21.3 & 29.3 & 33 \\
        SSB~\cite{patrick2020support} & None & 8.7 & 23.0 & 31.1 & 31.0 \\ 
        ActBERT~\cite{zhu2020actbert} & MSRVTT & 8.6 & \textbf{23.4} & \textbf{33.1} & 36 \\ 
        MIL-NCE~\cite{miech2020end} & None & 8.2 & 21.5	& 29.5 & 40 \\ 
        \midrule
        VT-TWINS & None & \textbf{9.4} & \textbf{23.4} & 31.6 & \textbf{32} \\
        \bottomrule
    \end{tabular}
    \end{adjustbox}
    \caption{\textbf{Text-to-Video Retrieval on MSRVTT.}}
    \label{tab:text2video_msrvtt}
\end{table}
\begin{table}[t!]
    \centering
    \setlength{\tabcolsep}{3.5pt}
    \begin{adjustbox}{width=0.47\textwidth}
    \begin{tabular}{l|c c c c|c c c c}
        \toprule
        \multicolumn{1}{c|}{\textbf{Method}} & \multicolumn{4}{c|}{\textbf{YouCook2}} & \multicolumn{4}{c}{\textbf{MSRVTT}} \\
        & R@1 & R@5 & R@10 & MedR & R@1 & R@5 & R@10 & MedR \\
        \midrule
        \midrule
        Random Init & 0.03 & 0.13 & 0.26 & 1717.5 & 0.1 & 0.49 & 0.98 & 499.5 \\
        MIL-NCE*~\cite{miech2020end} & 9.35 & 26.22 & 37.36 & 22 & 8.9 & 20.65 & 27.2 & 46 \\ 
        \midrule
        VT-TWINS & \textbf{9.7} & \textbf{28} & \textbf{40.3} & \textbf{16} & \textbf{9.1} & \textbf{22.9} & \textbf{29.1} & \textbf{43} \\
        \bottomrule
    \end{tabular}
    \end{adjustbox}
    \caption{\textbf{Video-to-Text retrieval.}
    * is our reproduction of official code of the MIL-NCE.
    }
    \label{tab:video2text}
\end{table}
\begin{table}[t!]
    \centering
    \begin{tabular}{l|l|c c c c}
        \toprule
        \multicolumn{1}{c|}{\textbf{Method}} & \multicolumn{1}{c|}{\textbf{Labeled Dataset}} & CTR \\
        \midrule
        \midrule
        Alayrac \etal~\cite{alayrac2016unsupervised} & IM, K400 & 13.3 \\
        CrossTask~\cite{zhukov2019cross} & IM, K400 & 22.4 \\
        CrossTask~\cite{zhukov2019cross} & IM, K400, CT & 31.6 \\
        Miech \etal~\cite{miech2019howto100m} & IM, K400 & 33.6 \\
        DWSA~\cite{shen2021learning} & CT & 35.5 \\
        ActBERT~\cite{zhu2020actbert} & CT & 37.1 \\ 
        MIL-NCE~\cite{miech2020end} & None & 35.5 \\
        \midrule
        VT-TWINS & None & \textbf{40.7} \\
        \bottomrule
    \end{tabular}
    \caption{\textbf{Action Step Localization on CrossTask.}}
    \label{tab:crosstask}
\end{table}

\noindent\textbf{Video-to-text retrieval.} We also compare the performance of video-to-text retrievals with MIL-NCE. 
Table~\ref{tab:video2text} shows that our VT-TWINS outperforms MIL-NCE on both YouCook2 and MSR-VTT.
Note that MIL-NCE blindly and equally treats all the captions in a time window around a query clip as positives.
We believe that this assumption often does not hold and learning with the inaccurate clip-caption pairs may hinder learning representations to precisely associate clips and captions.

\subsubsection{Action Step Localization}
We also evaluate the representations learned by our method in the action step localization task on the CrossTask dataset.
We adopted the zero-shot evaluation suggested in  ~\cite{miech2019howto100m}.
Table~\ref{tab:crosstask} shows that VT-TWINS significantly outperforms baselines achieving an CrossTask average recall (CTR) of 40.7\%.
This surpasses MIL-NCE (35.5\%) and even the models that are trained on the CrossTask dataset such as DWSA (35.5\%) and ActBERT (37.1\%).

\begin{figure}
    \centering
        \includegraphics[width=0.47\textwidth]{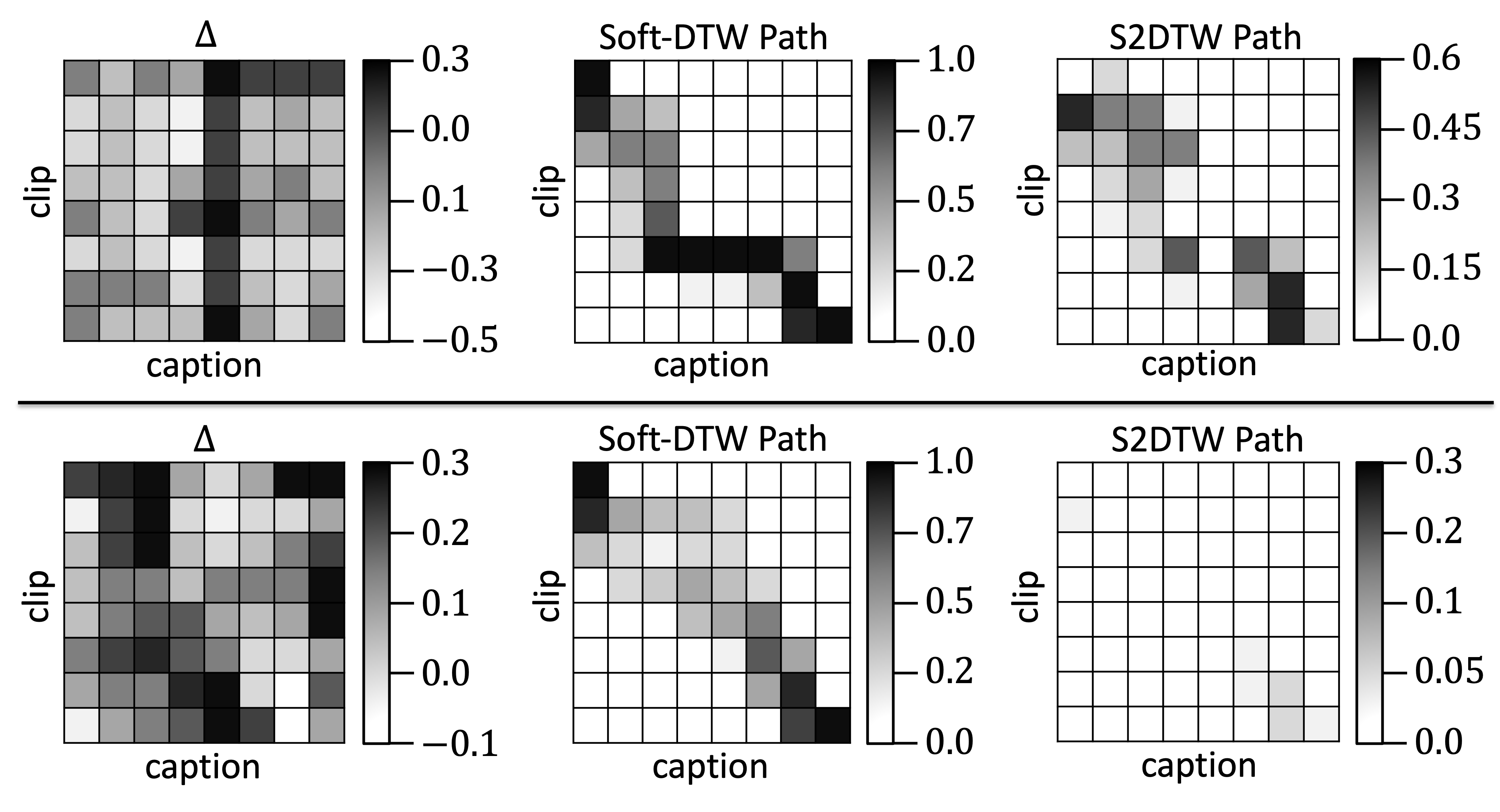}
    \caption{\textbf{Results of Weak Alignment.} 
    $\Delta$ is a pairwise distance matrix. The Soft-DTW path and the S2DTW path matrices are the gradient matrices $M$ and $\hat{M}$ defined in Section~\ref{subsec:smoothing}.
    Each row shows the partially and entirely irrelevant pairs, respectively.
    }
    \label{fig:skip}
\end{figure}

\subsection{Ablation Study and Qualitative Analysis}
\label{subsec:ablation_qualitative}

\subsubsection{Temporal Data Augmentation}
\label{subsubsec:augmentation}
As explained in Section 4.3, the proposed augmentation less likely generates a permutation which is significantly different from the original sequence.
To evaluate the effectiveness of our temporal data augmentation, we compare it with two other strategies: One is sampling from the uniform distribution and the other is sampling from a inverse distribution of our one, \ie, assigning a higher probability to the semantically similar permutation with the original sequence.
(2), (3), and (4) in Table~\ref{tab:ablation} demonstrate that temporal shuffles while maintaining semantic information helps to learn feature representation on weakly correlated data with non-sequential alignments.
Especially, the gap is substantial in the task that uses joint embedding representations (YouCook2, MSR-VTT, and CrossTask in Table~\ref{tab:ablation}) because strong augmentation harms semantic information a lot, it is difficult to learn the representations aligned between clip-captions.

\subsubsection{Weak Alignment}
\label{subsubec:skip}
The top row of Figure~\ref{fig:skip} shows the case of partially irrelevant pairs; the pairwise distance matrix $\Delta$ on top shows that the fifth caption has a consistently large distance from the other clips\footnote{Also refer to Figure~\ref{fig:partial} as an illustrative example.}.
In this case, the Soft-DTW is enforced to align one or more pairs per each timestamp. On the other hand, S2DTW shows the results that the unrelated pairs are weakly aligned because S2DTW skips them appropriately.
Moreover, the Soft-DTW has another problem that it is forced to align the start point (1,1) and the end point (n,m).
Unlike the Soft-DTW, we observe that S2DTW can ignore the start point and the end point.

The Soft-DTW also finds a temporal alignment path even in the entirely uncorrelated data like the case of Figure~\ref{fig:entire}.
The bottom row of Figure~\ref{fig:skip} illustrates that most elements of the pairwise distance are greater than zero (the leftmost matrix), \ie, the clips and captions are almost entirely irrelevant.
The path is clearly drawn in the Soft-DTW while most elements are not learned in S2DTW by aligning weakly.
(2) and (5) of Table~\ref{tab:ablation} show that weak alignment of S2DTW improves the performance on the weakly correlated data by ignoring irrelevant pairs.

\subsubsection{Local Neighborhood Smoothing}
\label{subsubec:smoothing}
We also evaluate the effectiveness of local neighborhood smoothing.
As mentioned in Section~\ref{subsec:smoothing}, local neighborhood smoothing can reflect local optimal path as well as global optimal path.
(5) and (6) in Table~\ref{tab:ablation} show that local neighborhood smoothing complements the DTW and improves the performance.

\begin{table}[t!]
    \centering
    \setlength{\tabcolsep}{3.5pt}
    \begin{tabular}{c|c|c|c||c c|c c|c}
        \toprule
        & \multicolumn{1}{c|}{\textbf{TA}} & \multicolumn{1}{c|}{\textbf{WA}} & \multicolumn{1}{c||}{\textbf{LS}} & \multicolumn{1}{c}{\textbf{HMDB}} & \multicolumn{1}{c|}{\textbf{UCF}} & \multicolumn{1}{c}{\textbf{YC2}} & \multicolumn{1}{c|}{\textbf{MV}} & \multicolumn{1}{c}{\textbf{CT}} \\
        \midrule
        \midrule
        (1) & - & - & - & 38.9 & 68.6 & 8.7 & 12.7 & 22.9 \\
        \midrule
        (2) & \textbf{A} & - & - & 39.4 & 69.3 & 9.6 & 13.6 & 23.5 \\
        (3) & \textbf{B} & - & - & 36 & 68.5 & 5 & 10.5 & 17.4 \\
        (4) & \textbf{C} & - & - & 36.9 & 68 & 4.9 & 11.5 & 16.8 \\
        \midrule
        (5) & \textbf{A} & \vv & - & 39.1 & 70.6 & 10.6 & 14.7 & 26.9 \\
        \midrule
        (6) & \textbf{A} & \vv & \vv & 42 & 72.1 & 12.5 & 17.4 & 28.2 \\
        \bottomrule
    \end{tabular}
    \caption{\textbf{Ablation Studies.}
    We report accuracy on the \textbf{HMDB} and \textbf{UCF}, R@10 on the YouCook2 (\textbf{YC2}) and MSR-VTT (\textbf{MV}), and CTR on the CrossTask (\textbf{CT}) to evaluate the contribution of the followings: temporal data augmentation (\textbf{TA}), weak alignment (\textbf{WA}), and local neighborhood smoothing (\textbf{LS}).
    Each element is applied to the standard DTW with a contrastive learning scheme.
    For \textbf{TA}, we evaluate the following strategies: \textbf{A}: suppressing semantically strong permutation (ours), \textbf{B}: random permutation, and \textbf{C}: encouraging semantically strong permutation (inverse of ours).
    } 
    \vspace{-5mm}
    \label{tab:ablation}
\end{table}
\section{Conclusion}
\label{sec:conclusion}
We have presented a novel multi-modal self-supervised learning framework for learning joint embeddings of video and text from uncurated narrated videos.
To address the challenges of weakly correlated video and caption pairs, our framework VT-TWINS first aligns the clips and captions by the proposed weak alignment algorithm and learns representations via contrastive learning.
Our experiments on a wide range of three tasks over five benchmark datasets demonstrate that the proposed method significantly improves the generality of joint embeddings and outperforms self-supervised methods as well as fine-tuned models on target tasks.
The proposed framework is a generic framework that is applicable in representation learning with multi-modal time-series data.
Future directions, limitations, and negative societal impacts are discussed in the appendix.

\vspace{4mm}
\noindent \textbf{Acknowledgments.}
This work was partly supported by Efficient Meta-Learning Based Training Method and Multipurpose Multi-Modal Artificial Neural Network for Drone AI (No.2021-0-02312), and ICT Creative Consilience program (IITP-2022-2020-0-01819) supervised by the IITP; the National Supercomputing Center with supercomputing resources including technical support (KSC-2021-CRE-0299) and Kakao Brain corporation. \newline
{\small
\bibliographystyle{ieee_fullname}
\bibliography{main}
}

\end{document}